\documentclass[letterpaper, 10 pt, conference]{ieeeconf}  

\IEEEoverridecommandlockouts                              
\usepackage{times} 
\usepackage{amsmath} 
\usepackage{amssymb}  

\usepackage[utf8]{inputenc}
\usepackage{hyperref}
\usepackage{booktabs}
\usepackage{multirow}
\usepackage{subcaption} 
\usepackage[skip=0pt,font=small,labelfont=bf]{caption}
\usepackage{array}
\usepackage{xspace}
\usepackage{flushend}
\usepackage{algorithm}
\usepackage{algorithmic}
\usepackage{siunitx}

\hypersetup{urlcolor=blue, colorlinks=true}

\usepackage[colorinlistoftodos,prependcaption,textsize=tiny]{todonotes}

\newenvironment{flushitemize}{%
\begin{list}{$\bullet$}
   {\setlength{\leftmargin}{5pt}}%
    \setlength{\labelwidth}{10pt}
    \setlength{\itemindent}{0pt}
    \setlength{\labelsep}{0.5em}
 \setlength{\itemsep}{1pt}
 \setlength{\parskip}{0pt} 
 \setlength{\parsep}{0pt}}
 {\end{list}}

\newcommand{\structformer}{\emph{StructFormer}}
 
\begin{document}
\title{StructFormer: Learning Spatial Structure\\ for Language-Guided Semantic Rearrangement of Novel Objects}

\author{Weiyu Liu\(^1\), Chris Paxton\(^2\), Tucker Hermans\(^{2,3}\), and Dieter Fox\(^{2,4}\)\thanks{\(^1\)Georgia Tech. \(^2\)NVIDIA. \(^3\)Univ. of Utah. \(^4\)Univ. of Washington}}

\input{intro-figure}
\maketitle

\begin{abstract}
  Geometric organization of objects into semantically meaningful arrangements pervades the built world. As such, assistive robots operating in warehouses, offices, and homes would greatly benefit from the ability to recognize and rearrange objects into these semantically meaningful structures.
  To be useful, these robots must contend with previously unseen objects and receive instructions without significant programming. While previous works have examined recognizing pairwise semantic relations and sequential manipulation to change these simple relations none have shown the ability to arrange objects into complex structures such as circles or table settings.
To address this problem we propose a novel transformer-based neural network, \structformer{}, which takes as input a partial-view point cloud of the current object arrangement and a structured language command encoding the desired object configuration. 
  We show through rigorous experiments that \structformer{} enables a physical robot to rearrange novel objects into semantically meaningful structures with multi-object relational constraints inferred from the language command. 
\end{abstract}

\section{Introduction}\label{sec:intro}
Organizing objects into complex, semantically meaningful arrangements has many real-world applications, such as setting the table, organizing books, and loading the dishwasher.
Indeed, the organizing task received the second most internet searches on the comprehensive chore list from~\cite{Cakmak2013TowardsAC} suggesting its practical importance for future domestic robots. More generally rearrangement was recently proposed as a benchmark for embodied AI~\cite{Batra2020RearrangementAC}. However, to be broadly deployed these robots should ideally receive instructions in a way that doesn't overly burden the operator assigning the task. In this work, we focus on the problem of semantic rearrangement, where a robot must move a set of novel objects to form a spatial structure that satisfies a high-level language instruction such as \textit{put the mugs in a row}, \textit{build a circle}, and \textit{set the table} as shown in Fig.~1.

Successful rearrangement manipulation in unstructured environments requires a robot to jointly reason about geometric and semantic properties of these novel objects and the objects' physical interactions~\cite{Batra2020RearrangementAC,paxton2021predicting}.
Realistic settings additionally present objects not required to accomplish the task requiring the robot to reason about additional relations between structures of interest and distractor objects.
Finally, the robot needs an encoding of its goal compatible with efficient planning, while also being easily provided by the operator.
Language provides an obvious input modality for untrained users to specify goals~\cite{Tellex2020RobotsTU}; however, it brings with it challenges in inferring the implied object configuration and generating representations interpretable by the robot. 

Previously, researchers have worked on modeling, interpreting, and grounding relations between scene elements~\cite{krishna2017visual, shridhar2020ingress,kunze2014combining,gunther2018context}. Such reasoning can enable robots to move a query object with respect to an anchor object to satisfy a desired spatial relations~\cite{mees2020learning,janner2018representation,paxton2021predicting,venkatesh2020spatial,kartmann2020representing,yan2020robotic}. More complex spatial structures from language instructions have be generated in a blocks world environment by chaining these binary relation changing manipulations~\cite{bisk2018learning}. Others have leveraged spatio-semantic relations between pairs of objects as an abstraction for planning \cite{zhu2020hierarchical,kase2020transferable,paxton2019prospection,zeng2018semantic}. Alternatively, robots can perform multi-object manipulation by directly learning to transform object groups without associated semantic reasoning~\cite{wilson-corl2019-collection-pushing}.

In contrast to the pairwise reasoning of prior work, our model explicitly reasons about multi-object semantic relations given a structured language instruction to specify the goal. We treat the semantic rearrangement planning problem as a sequential prediction task using a novel transformer-based architecture named \structformer{}.
We use transformer encoders to build a contextualized representation of abstract concepts expressed in language instructions as well as semantic and geometric multi-object relations. This enables \structformer{} to reason over a variety of objects of varying number without access to object models. The encoder can directly predict what objects to move and also provide a context for an autoregressive transformer decoder to predict where the objects should go. 

To train and validate our approach, we procedurally generate a dataset of four different structures---circles, lines, towers, and table settings---using models from 335 real-world objects. 
%
We tested our method both on novel objects and arrangements in simulation, and on a physical robot.
Our experiments show that \structformer{}'s ability to directly reason over multiple objects enables higher manipulation success compared to a model that only reasons about pairwise relations between objects.
For further videos, gifs, and supplemental material please visit our website \url{https://sites.google.com/view/structformer}.

\section{Related Work}\label{sec:related_work}
\noindent\textbf{Grounding spatial relations} defines prior work on spatial understanding focusing on modeling relations between pairs of objects, such as \emph{in}, \emph{on}, and \emph{left of}~\cite{rosman2011learning,fichtl2014learning,mees2017metric,yuan2021sornet}. These spatial relations have been used to specify action goals (e.g., moving the cup to the right of the bowl) \cite{mees2020learning,janner2018representation}. More complex structures such as towers or letters can also be created by chaining these action goals, as shown in \cite{bisk2018learning}, while \cite{Paul2016EfficientGO} grounds such spatial abstractions. Other works examine classifying spatial structures~\cite{teodorescu2020spatialsim}.
Besides facilitating communicating language goals, spatial relations have also been used as an abstraction for planning \cite{zhu2020hierarchical}. A recent work segments video demonstrations of object manipulations into sequences of spatial relations, demonstrating the benefit of this abstract representation for imitation learning \cite{hristov2020disentangled}.  
Different from existing methods, we model spatial relations between multiple objects, extending from binary spatial relations to multi-object relations.

\noindent\textbf{Visual question answering and understanding} reasons about spatial and compositional structures of visual elements--an active research area in the vision community~\cite{johnson2017clevr,yi2018neural,ding2020object,nazarczuk2020shop}. The CLEVR visual question answering (VQA) dataset \cite{johnson2017clevr} and subsequent extensions \cite{girdhar2020cater,hong2021transformation,CLEVRER2020ICLR} have provided useful benchmarks for developing systems that perform object-centric reasoning. More recent efforts leverage RAVEN's Progressive Matrices to test models' abilities to discover structure among objects \cite{zhang2019raven,barrett2018measuring}. The NLVR dataset also helps develop methods to explicitly reason about sets of objects \cite{suhr2017corpus}. In contrast to passively parsing objects in the scene, our method actively manipulates objects to achieve desired structures. 

\begin{figure*}[t]
  \centering
  \includegraphics[width=0.99\textwidth]{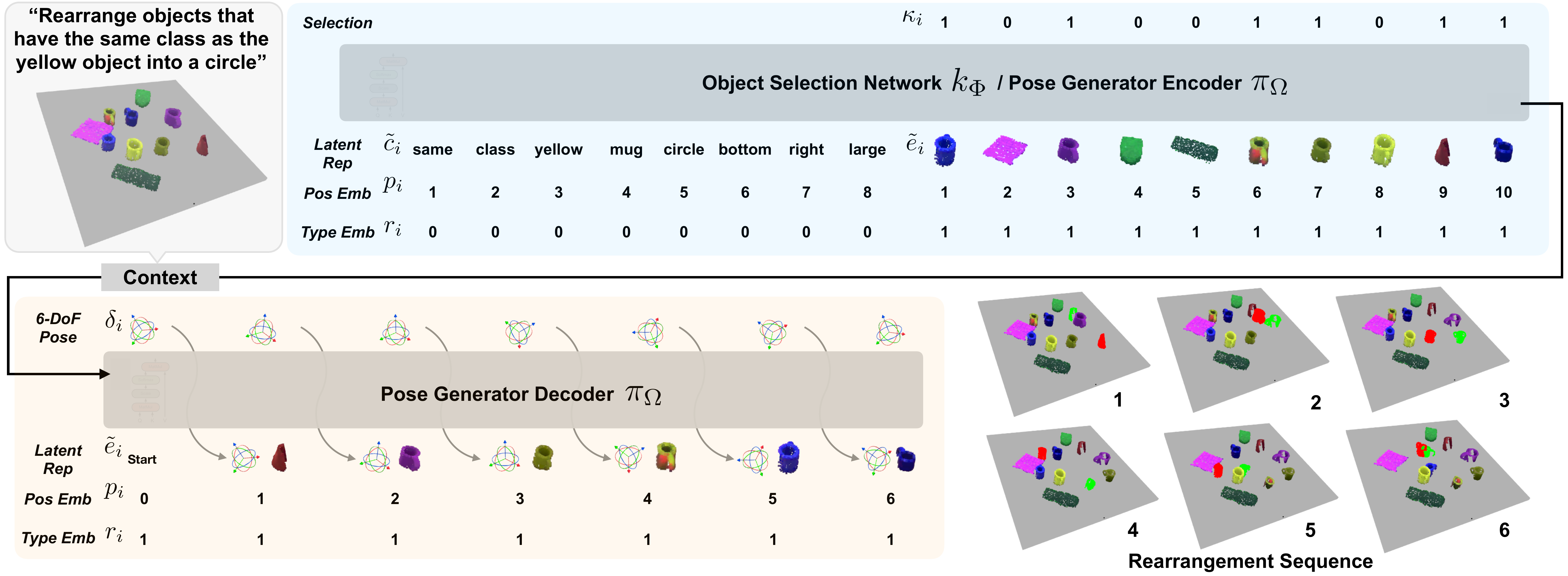}
  \caption{Visualization of the components of \structformer{}. \structformer{} takes as input a sequence of object point clouds and language instruction, passes this through an object selection network and pose generator to output a sequence of object rearrangement actions to achieve the configuration implied by the language command. We describe each component in detail in Sec.~\ref{sec:approach}}
  \label{fig:system}
  \vspace{-.5cm}
\end{figure*}

\noindent\textbf{3D structure synthesis} 
methods can be categorized into holistic generative models~~\cite{achlioptas2018learning,park2019deepsdf}, which directly generate the whole 3D structure, and structure-aware models \cite{mo2019structurenet,wu2020pq,li2017grass,li2019grains,Chaeibakhsh-icores2021-room-design,jiang2012learning}, which leverages substructures (e.g., object parts or furniture in a room). Our work is most closely related to the latter. These models decompose generations into synthesizing object parts and creating spatial arrangements of parts. Instead of leveraging structure prior of object parts (e.g., a table has four legs on each corner), we condition spatial arrangements of objects based on high-level language instruction.


\section{Transformer Preliminary}\label{sec:transformer}
Transformers were proposed in \cite{vaswani2017attention} for modeling sequential data. At the heart of the Transformer architecture is the scaled dot-product attention function, which allows elements in a sequence to attend to other elements. Specifically, an attention function takes in an input sequence $\{x_1,...,x_n\}$ and outputs a sequence of the same length $\{y_1,...,y_n\}$. Each input $x_i$ is linearly projected to a query $q_i$, key $k_i$, and value $v_i$. The output $y_i$ is computed as a weighted sum of the values, where the weight assigned to each value is based on the compatibility of the query with the corresponding key. The function is computed on a set of queries simultaneous with matrix multiplication.
\begin{align}
    \text{Attention}(Q,K,V) = \text{softmax}(\frac{QK^T}{\sqrt{d_k}})V
\end{align}
where $d_k$ is the dimension of queries and keys. The queries, keys, and values are stacked together into matrix $Q \in \mathbb{R}^{n \times d_{\text{model}}}$, $K \in \mathbb{R}^{n \times d_{\text{model}}}$, and $V \in \mathbb{R}^{n \times d_{\text{model}}}$. The original transformer architecture has a stack of encoder layers and a stack of decoder layers. Each encoder layer includes an attention layer and a position-wise fully connected feed forward network. Each decoder layer has one additional attention layer, which performs attention over the output of the encoder stack. With the causal attention mask, the decoder layer can prevent the predictions from relying on future inputs, therefore suitable for autoregressive generation. We refer the readers to the original paper for details~\cite{vaswani2017attention}.


\section{\structformer{} for Object Rearrangement}\label{sec:approach}
Given a single view of a scene $s$ containing objects $\{o_1,...,o_N\}$ and a structured language instruction $l$ containing word tokens $\{w_1,...,w_M\}$, our goal is to rearrange a subset of the objects $\{o_1,...,o_{N_q}\}$, which we call \textit{query objects}, to reach a goal scene $s^*$. The rearranged scene should satisfy the spatial and semantic constraints encoded in the instruction $l$ while being physically valid. We assume we are given a partial-view point cloud of the scene $Z$ with segment labels for points to identify the different objects. Given this point cloud $Z$ and the language instruction $l$, the robot must find pose offsets $\{\delta_1,...,\delta_{N_q}\}$, which can transform the query objects $\{o_1,...,o_{N_q}\}$ in the initial scene to new poses that satisfy the goal arrangement implied by \(l\).

Our model consists of an object selection network and a pose generator network, as shown in Fig.~\ref{fig:system}. Both networks jointly reason about the object point clouds and language instructions using transformers~\cite{vaswani2017attention}. For a given scene, latent representations of words and objects are used to construct the input sequence $\{w_1,..,w_N, o_1,...,o_N\}$ for the two networks. The object selection network uses a transformer encoder to predict output sequence $\{\kappa_1,...\kappa_N\}$ in a single forward inference, where $\kappa_i$ is a binary variable indicating if the robot should rearrange object $o_i$. The pose generator network uses a transformer decoder to autoregressively generate a sequence of pose offsets $\{\delta_1,...,\delta_{N_q}\}$, which we achieve by feeding in representations of the selected objects $\{o_1,...,o_{N_q}\} = \{o_i |\kappa_i = 1\}_{i=1}^N$ as targets for decoding. Below, we discuss how we build our latent representations of objects and language instructions. Then we describe the two transformer networks in detail. Finally, we discuss how to train our system and use it for rearrangement planning.

\subsection{Object and Sentence Encoders}
To jointly model objects and language instructions, we convert point clouds and word tokens to hidden representations. Given the segmented point cloud $x_i$ of an object $o_i$, we learn a mapping $h_o(x_i) \rightarrow \Tilde{e}_i$, in order to obtain the latent representation of the object. We base our object encoder around the point cloud transformer (PCT) model~\cite{guo2021pct}, which leverages the transformer as a permutation-invariant function for learning from unordered points. We also map each unique word token from the language instructions to an embedding with a learned mapping $h_w(w_i) \rightarrow \Tilde{c}_i$. We use a learned position embedding $h_{pos}(i) \rightarrow p_i$ to indicate the position of the words and objects in input sequences and a learned type embedding $h_{type}(\tau_i) \rightarrow r_i$ to differentiate object point clouds ($\tau=1$) and word tokens ($\tau=0$). We concatenate our latent codes together to obtain the final object $e_i = [\Tilde{e}_i; p_i; r_i] $ and word $c_i = [\Tilde{c}_i; p_i; r_i]$ embeddings.

\subsection{Object Selection Network}
The object selection network $k_{\Phi}(\{e_i\}, \{c_i\}) \rightarrow \{\kappa_i\}$ predicts objects that need to be rearranged based on the language instruction. We use a transformer encoder to perform relational inference over all objects in a scene and the given language instruction (e.g., identifying ``objects that are smaller than a pan''). The object and sentence encoders encode the words $\{w_i\}$ and objects $\{o_i\}$ to create the input sequence $\{c_1,..,c_M, e_1,...e_N\}$ to the transformer encoder. We feed encoder's output at each object's position into a linear layer to predict $\{\kappa_1,...\kappa_N\}$. Formally, the object selection transformer models the distribution
\[p(\{\kappa_i\}_{i=1}^N|\{e_i\}_{i=1}^N, \{c_i\}_{i=1}^N) \approx \prod_{i=1}^{N} p(\kappa_i|\{e_i\}_{i=1}^N, \{c_i\}_{i=1}^N)\]

\subsection{Language Conditioned Pose Generator}
We learn a generative distribution $\pi_{\Omega}(\{e_i\},\{c_i\}) \rightarrow \{\delta_i\}$ over possible pose offsets for objects that might satisfy the language instruction and are physically valid. We use a transformer encoder-decoder model. The encoder has the same architecture as the object selection network and encodes the sequence $\{c_1,..,c_M, e_1,...e_N\}$ to build a contextualized representation of the language instruction and objects in the scene, including objects that need to be moved and objects that will remain stationary. The decoder autoregressively predicts each object's pose offset, conditioning on the global context and the pose offsets of previously predicted objects. Formally, the decoder takes as input the sequence $\{e_{0}, [\delta_{0}; e_1], [\delta_{1}; e_2],...,[\delta_{N_q - 1}; e_{N_q}]\}$ and predicts $\{\delta_{0}, \delta_{1},...,\delta_{N_q}\}$. We ensure the input object poses are not used by the decoder by shifting the input poses by one position and using a causal attention mask. We model the following distribution with the encoder-decoder model
\[p(\{ \delta_i \}_{i=0}^{N_q}|\{e_i\}_{i=1}^{N}, \{c_i\}_{i=1}^{M}) =  \prod_{n=0}^{N} p(\delta_i|\delta_{<i}, \{e_i\}_{i=1}^{N}, \{c_i\}_{i=1}^{M})\]
The network obtains its stochasticity by using a dropout layer with probability $p \in [0, 1]$ during training and inference. 

We parameterize 6-DoF pose offset $\delta$ as $(t, R) \in SE(3)$. We directly predict $t \in \mathbb{R}^3$ and predict two vectors $a, b \in \mathbb{R}^3$, which are used to construct the rotation matrix $R \in SO(3)$ using a Gram–Schmidt-like process proposed in \cite{zhou2019continuity}. In contrast to quaternion and eular angle representation, this representation has no discontinuities and facilitates learning accurate 6-DoF placement poses. $\delta_{0}$ is the pose of a virtual structure frame in the world frame. $\delta_{i}, \forall i>0$ defines the 3D position of object $o_i$ in the structure frame and the relative rotational offset between its target and initial pose. $e_0$ is a learned embedding for the structure frame. 

The order of the query objects in the input sequence is predefined for each spatial structure. For example, the rearranged objects will build a circle structure clockwise. We find empirically that imposing an order on objects and using a virtual frame help create precise spatial structures.


\subsection{Inference and Training}
During inference, we select objects to rearrange based on prediction from the object selection network $k_\Phi$. We sample a batch of $B$ rearrangements from our pose generator $\pi_\Omega$ for the query objects. For each sample, we autoregressively predict the target pose of the structure frame and each object, conditioned on the previous predictions in the sequence. 

We train the object selection network $k_\Phi$ and pose generator $\pi_\Omega$ with data from rearrangement sequences. The object selection network is trained on initial scenes and groundtruth query objects using a binary cross entropy loss. The generator is trained with an $L2$-loss minimizing the distance between groundtruth and predicted placement poses.

\section{Data Generation}\label{sec:data}
We introduce a dataset containing more than 100,000 rearrangement sequences. We pair each rearrangement with a high level language instruction specifying the target spatial rearrangement for a set of objects. The language instructions involve many different semantic and geometric properties for both grounding objects and specifying the spatial structures, as shown in Table~\ref{tab:concepts}. We procedurally generate stable and collision-free object arrangements in the PyBullet physics simulator~\cite{coumans2017pybullet} and render with the photo-realistic image render NVISII~\cite{morrical2021nvisii}.
    \begin{table}[t]
        \centering
        \caption{Semantic and spatial concepts in our dataset.}
        \resizebox{\columnwidth}{!}{
        \begin{tabular}{lll}
        \toprule
        Entity & Type (\# Value)  & Values \\
        \midrule
        \multirow[t]{5}*{obj} &class (35) & basket, beer bottle, book, bowl, calculator, \\&& candle, controller, cup, donut, ... \\
        &material (3) & glass, metal, plastic \\
        &color (6)& blue, cyan, green, magenta, red, yellow \\
        &relate (3) & less, equal, more \\
        \midrule
        \multirow[t]{5}*{struct}&shape (4) & circle, line, tower, table setting\\
        &size (3) & small, medium, large\\
        &vertical position (3) & top, middle, bottom \\
        &horizontal position (3) & left, center, right \\
        &rotation (4) & north, east, south, west \\  
        \bottomrule
        \end{tabular}}
        \label{tab:concepts}
        \vspace{-.3cm}
    \end{table}
\begin{figure}[t]
  \centering
  \includegraphics[width=0.48\textwidth]{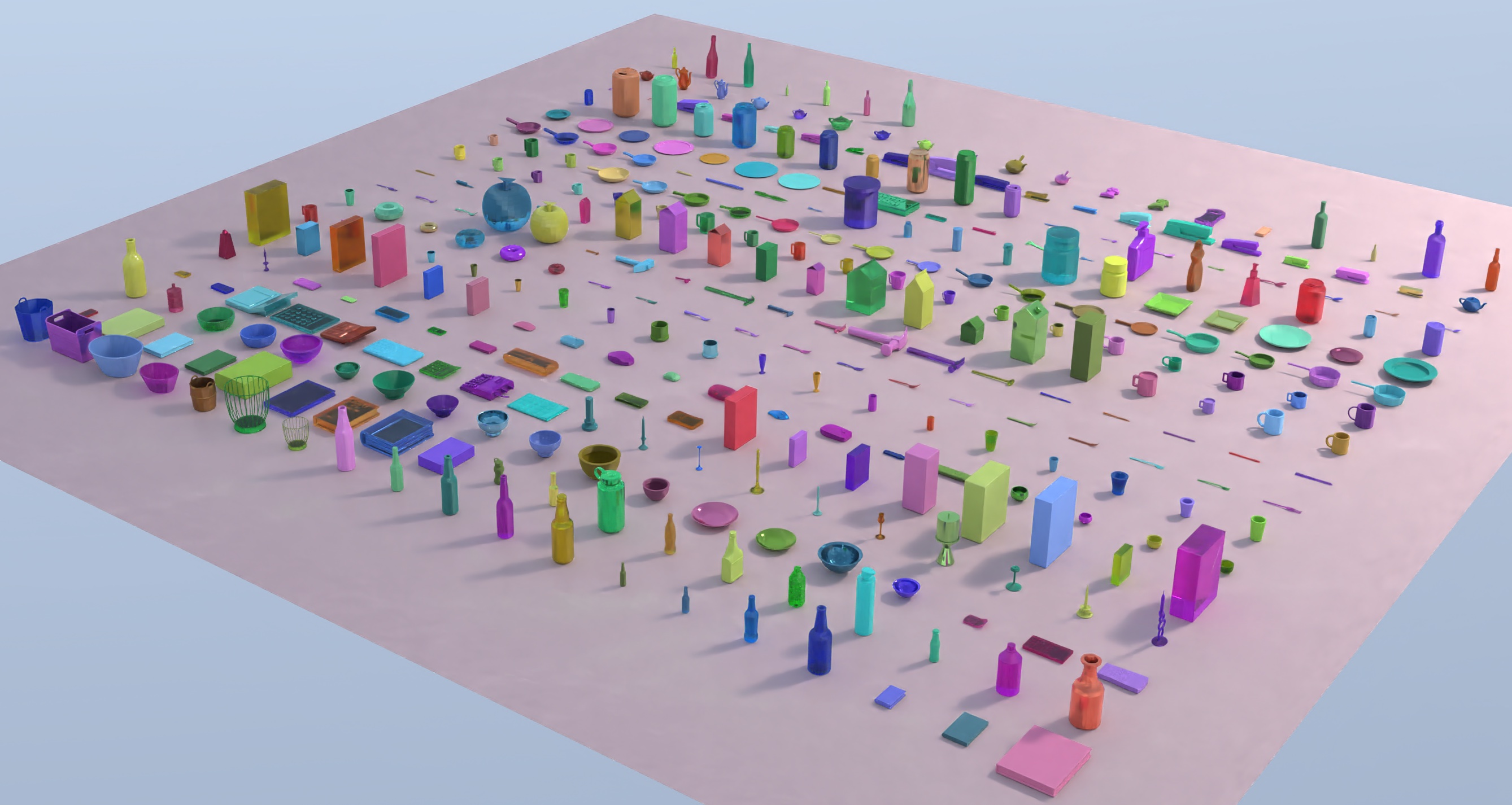}
  \caption{Objects used in generating our arrangement dataset organized by the 35 object classes.}
  \label{fig:objects}
  \vspace{-24pt}
\end{figure}
With the goal of generalization in mind, we adopt 335 everyday household objects from the acronym dataset~\cite{acronym2020}. Figure~\ref{fig:objects} shows the diversity of objects used from 35 distinct classes.

We generate a language-conditioned rearrangement sequence in three steps: (1) sampling a referring expression for query objects, (2) arranging query objects into a physically realistic spatial structure, and (3) creating an action sequence with time reversal. We discuss each step in detail below.

We functionally generate referring expressions for sets of objects that need to be rearranged. A referring expression can indicate the query objects explicitly with a discrete feature (e.g., metal objects) or by relating to an anchor object, which in turn can be described by one to three discrete features.  Using anchor objects allows us to create referring expressions that require relational reasoning of abstract semantic properties of objects (e.g., objects that have the same material as the blue bottle) and continuous geometric properties (e.g., objects that are shorter than the glass cup). After sampling a referring expression, we add query objects, an optional anchor object, and additional distractor objects, which do not match the referring expression, to the scene. Since table settings involve specific objects, we do no create referring expressions for this structure type.
\begin{figure}[thb]
  \centering
  \includegraphics[width=0.49\textwidth]{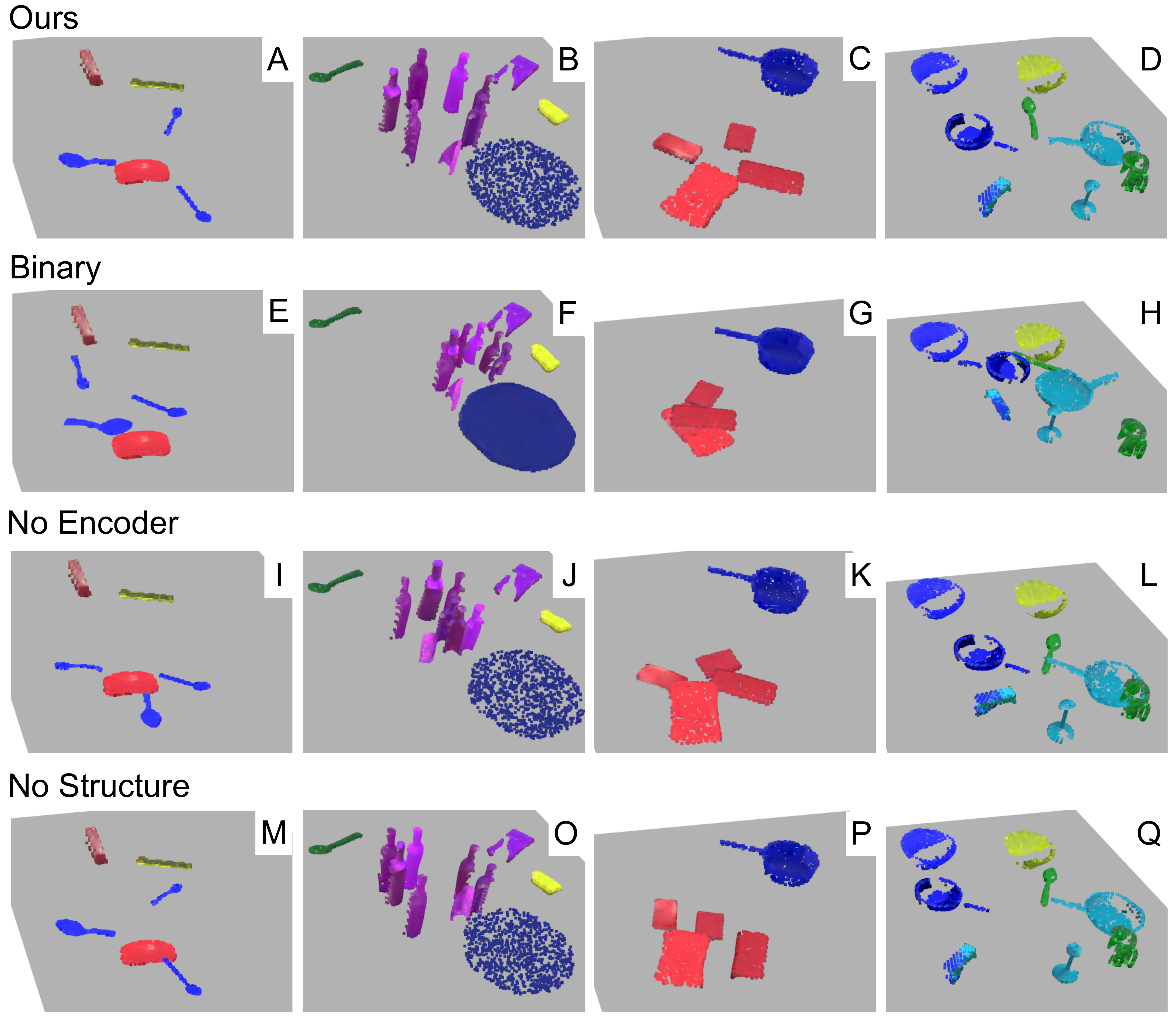}
  \caption{Visualizations of the predicted circular arrangements from our pose generator and baselines. Our model creates precise structures for different numbers of objects with various geometries.}
  \label{fig:prior_comparison}
  \vspace{-.6cm}
\end{figure}

We rearrange query objects into physically correct instances of one of the four defined spatial structures according to different geometric parameters (e.g., radius, size, position, rotation) in PyBullet.
By discretizing the parameters of the structure according to a pre-specified vocabulary, we generate a sentence describing the structure (e.g., place query objects into a large circle on the top right of the table).

Finally, we move objects out of the structure to random, collision-free poses in the scene. We obtain an action sequence that rearranges objects from random poses into the goal configuration described by the language instruction by reversing the random action and associated image sequences. We use NVISII to render color and depth images and instance segmentation masks of all objects in the sequence. 



\section{Experiments}\label{sec:experiments}

\begin{table*}[t]
  \centering
  \caption{Our pose generator produces smaller translational and rotational errors of predicted object poses for four structures.}
  \begin{tabular}{lcccccccccccccccc}
    \toprule
    \multirow{3}{*}{Model} & \multicolumn{4}{c}{Circle} & \multicolumn{4}{c}{Line} & \multicolumn{4}{c}{Tower} & \multicolumn{4}{c}{Table Setting} \\  
    \cmidrule(lr){2-5}\cmidrule(lr){6-9}\cmidrule(lr){10-13}\cmidrule(lr){14-17}
                           & \multicolumn{2}{c}{Obj} & \multicolumn{2}{c}{Struct} & \multicolumn{2}{c}{Obj} & \multicolumn{2}{c}{Struct} & \multicolumn{2}{c}{Obj} & \multicolumn{2}{c}{Struct} & \multicolumn{2}{c}{Obj} & \multicolumn{2}{c}{Struct}\\
    \cmidrule(lr){2-3}\cmidrule(lr){4-5}\cmidrule(lr){6-7}\cmidrule(lr){8-9}\cmidrule(lr){10-11}\cmidrule(lr){12-13}\cmidrule(lr){14-15}\cmidrule(lr){16-17}
                           & $t$ & $R$ & $t$ & $R$& $t$ & $R$& $t$ & $R$& $t$ & $R$& $t$ & $R$& $t$ & $R$ & $t$ & $R$ \\
    \midrule
    Binary &11.40&76.36&/&/&5.64&58.32&/&/&6.35&65.62&/&/&7.87&46.91&/&/ \\
    No Structure &8.32&53.58&/&/&4.24&40.26&/&/&6.12&74.02&/&/&4.79&46.22&/&/\\
    No Encoder &5.43&54.52&8.85&\textbf{22.47}&4.19&45.39&5.97&\textbf{0.00}&2.86&41.14&\textbf{10.58}&22.63&3.35&46.57&6.08&\textbf{0.01}\\
    Ours &\textbf{3.54}&\textbf{42.08}&\textbf{8.60}&22.55&\textbf{2.95}&\textbf{39.66}&\textbf{5.86}&0.11&\textbf{2.81}&\textbf{37.17}&10.61&\textbf{22.55}&\textbf{3.09}&\textbf{42.43}&\textbf{5.33}&0.05 \\
    \bottomrule
    \vspace{-20pt}
  \end{tabular}
  \label{tab:prior_result}
\end{table*}

In this section, we provide rigorous experimental validation of our approach. We first evaluate the individual components of \structformer{} on the collected dataset. Following this we show planning performance for our entire system. We then provide results for using our system to generate and execute rearrangement plans on a physical robot with real-world objects and sensing.

\subsection{Model Component Testing}
We split the dataset into 80\% training, 10\% validation, and 10\% testing. The test data consist of new target spatial structures and novel object combinations.
\subsubsection{Language Conditioned Pose Generator}
We compare our pose generator to the following baselines.
\begin{flushitemize}
    \item \textbf{Binary}: This baseline uses a transformer encoder to encode input $\{w_1,...,w_M, x_j, Z_{\sim o_j}, x_{j+1}\}$ and predict $\delta_{j+1}$, where $x_{j+1}$ is the point cloud of the current object to move and $x_j$ is that of the previously moved object $o_j$, which serves as a spatial anchor. $Z_{\sim o_j}$ is the scene point cloud $0.5\mathrm{m}$ around $o_j$. To rearrange objects, we iteratively decode each object's placement. For the first object, $x_j$ is omitted and the entire scene point cloud $Z$ is used in place of $Z_{\sim o_j}$. This baseline allows us to investigate whether modeling pairwise relations between objects alone is enough for generating complex spatial structures.
    \item \textbf{No Encoder}: This variant of our pose generator does not use the transformer encoder to extract global context for decoding. When predicting the pose offset for an object, it only has information about the language instruction and previously predicted objects. This baseline is similar to the previous transformer models used in floor plan generation and clip-art generation work \cite{wang2020sceneformer,radevski2020decoding}, where the placements of objects are less spatially constrained.
    \item \textbf{No Structure}: This variant of our pose generator directly predicts 6-DoF pose offset of each object in the world frame without predicting and using the virtual structure frame.
\end{flushitemize}

We report average Euclidean distances between groundtruth and predicted positions $t \in \mathbb{R}^3$ in centimeters and geodesic distances between groundtruth and predicted rotation matrices $R \in SO(3)$ in degrees. 

As shown in Table~\ref{tab:prior_result}, our model outperforms all baselines at predicting precise placements of objects for four different structures. Comparison with \textbf{No Encoder} validates that the transformer encoder in our pose generator is crucial for precise rearrangements. The larger errors produced by the \textbf{No Struct} baseline indicate that using a virtual structure frame helps anchor placements of objects for better structure generation. This is analogous to leveraging one object as the spatial anchor for placing another object when manipulating pairwise spatial relations \cite{mees2020learning,paxton2021predicting}. Additionally, we hypothesize that separately predicting the placements of the structures and objects also helps deal with spatial ambiguities embedded in language instructions (e.g., arrange a circle in the \textit{middle} of the table). Finally, we highlight the performance difference between \textbf{Binary} and our model, which confirms that modeling multi-object spatial relations is beneficial not only for creating complex spatial structures but also for generating simpler structures such as lines and towers, which can be described by binary relations. The "Struct" results in Table~\ref{tab:prior_result} show that we have comparable accuracy in locating the target structure relative to anchor objects in the scene with or with out the use of the encoder.

Besides better quantitative performance achieved by our method, we also see qualitative improvement. In Fig.~\ref{fig:prior_comparison}, we visualize predicted rearrangements by transforming point clouds of objects. We highlight that the \textbf{No Encoder} and \textbf{Binary} baselines are inadequate at producing circular structures because these two methods are not able to condition placement of an object based on future objects (e.g., how many more objects to be placed and what are their dimensions). We also note that \textbf{No Struct} fails to predict precise placement for a large number of objects (O) and is inconsistent at producing accurate alignments of objects (P).

    \begin{figure}[t]
		\centering
		\includegraphics[width=0.48\textwidth]{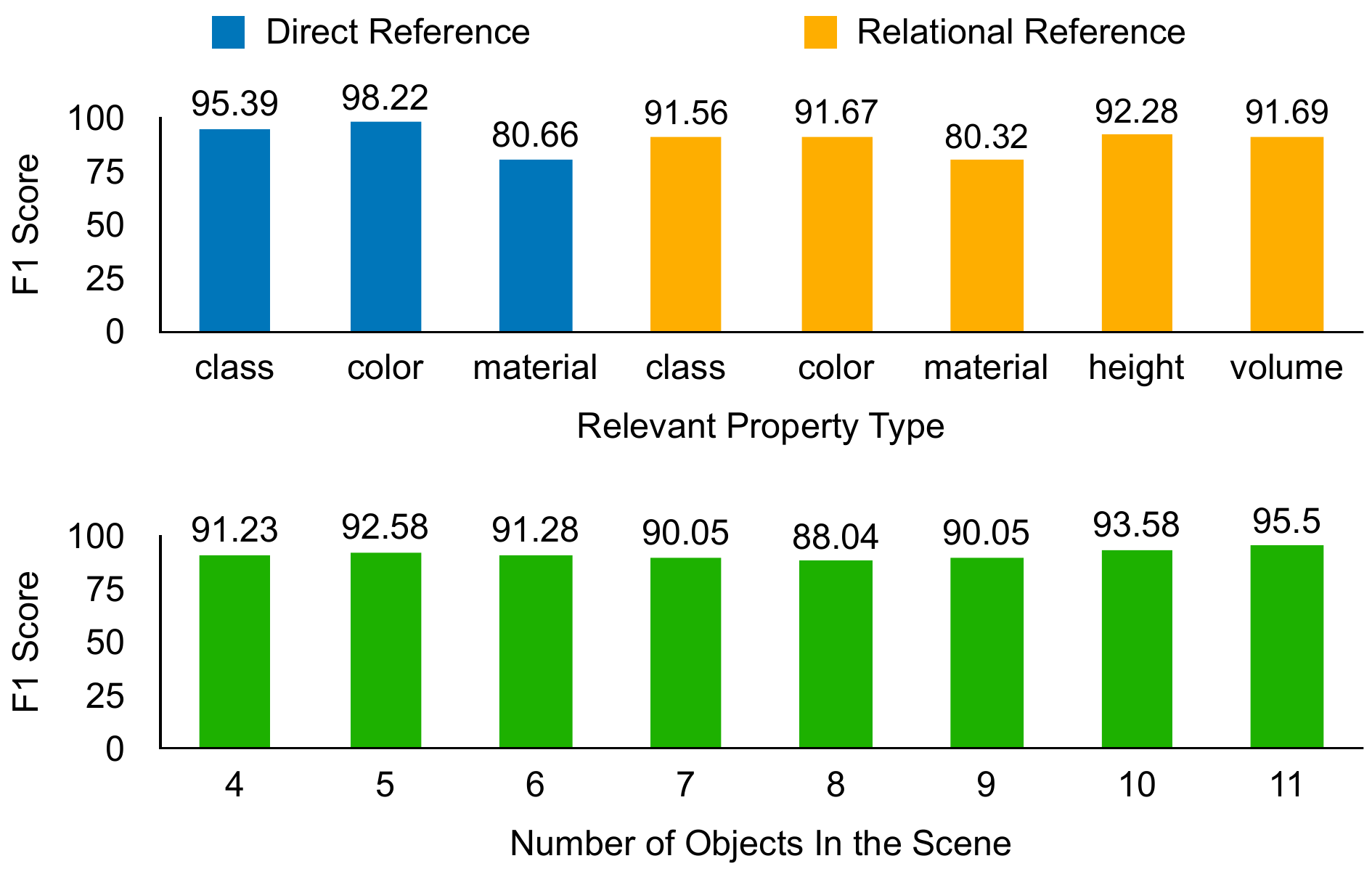}
		\caption{F1 scores for query objects predicted by our object selection network. The top chart organizes performance based on associated property types in referring expressions. The bottom orders results by the number of objects identified in referring expressions.}
		\label{fig:object_selection_results}
		\vspace{-.2cm}
    \end{figure}
    \begin{figure}[t]
		\centering
		\includegraphics[width=0.49\textwidth]{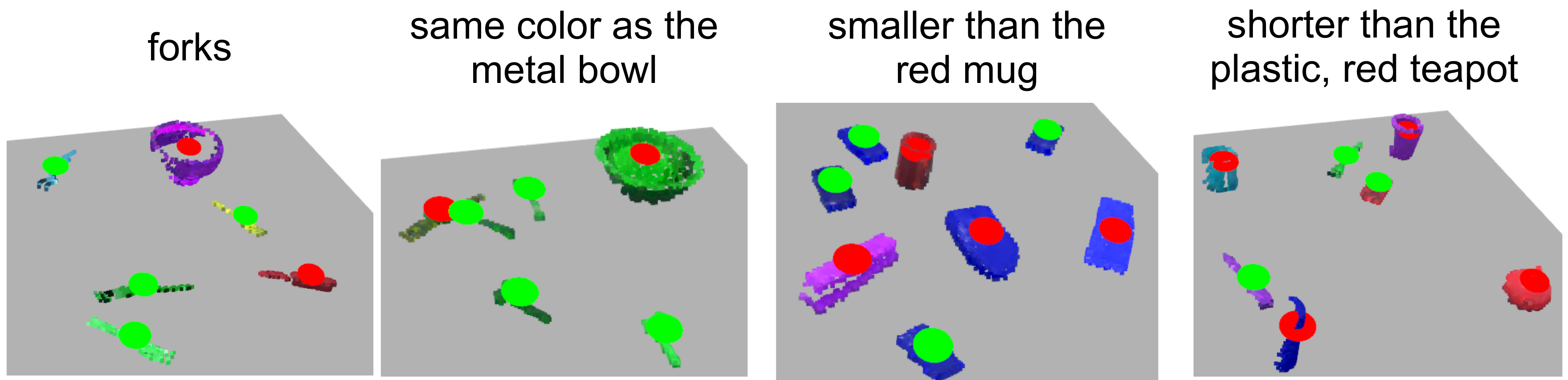}
		\caption{Examples of objects predicted to move based on referring expressions. Green dots indicate objects selected by our network.}
		\label{fig:object_selection_example}
		\vspace{-.5cm}
    \end{figure}

\subsubsection{Object Selection Network}
We test our object selection network on novel combinations of objects.  Fig.~\ref{fig:object_selection_results} shows that our model can retrieve objects based on both direct references and relational references and can also reason about continuous-valued properties including height and volume. Fig.~\ref{fig:object_selection_results} further demonstrates that our model maintains high accuracy when given an increasing number of target objects. We visualize identifying objects to move in Fig.~\ref{fig:object_selection_example}.

\subsection{Evaluating Full System in Simulation}
We evaluated our entire system in the simulation environment using 138 novel object models from 23 known object classes. We preserve physical interactions of objects while using ground-truth instance segmentation and omitting low-level control of the robot. We simulate object placement by dropping any moved object from $3$ cm above the predicted target $z$ value. For each scene, we use the same procedure as our data generation to sample a referring expression and corresponding query, anchor, and distractor objects. After putting objects into the scene, we randomly select a structure and its parameters to create a high-level language instruction.
    
    \begin{figure}[t]
		\centering
		\includegraphics[width=0.48\textwidth]{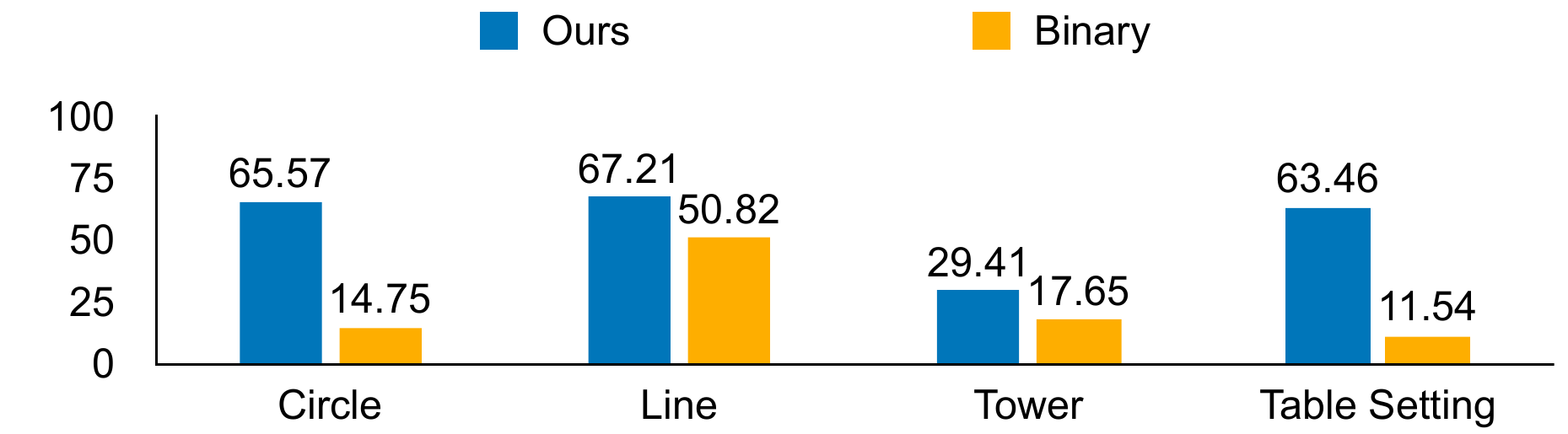}
		\caption{Success rate comparison to the Binary baseline for generating semantically correct, physically valid structures in simulation.}
		\label{fig:simulation_evaluation}
		\vspace{-20pt}
    \end{figure}
    
We first tested the combined system. In our experiments, the object selection network was able to identify all query objects given a referring expression in 95/156 (61\%) of the tested scenes and identify 70\% of the specified objects in 129/156 (83\%) of the scenes.  With the selected objects, the pose generator successfully rearranged objects into circles, lines, and towers in 40/61 (66\%), 40/61 (65\%) and 10/34 (29\%) of the scenes. The table setting structure was not included because it does not use the same object referring expressions as the other three structures. The overall success rate of the whole pipeline on building spatial structures based on high-level language instructions was 58/156 (37\%). Constructing tower structures was especially challenging because many tested objects have irregular shapes and inherently cannot be stacked (e.g., apples, teapots, candle stands). Another major failure mode was incompatible structure parameters and objects (e.g., arranging large pans into a small circle).

We also compared our pose generator to the \textbf{Binary} baseline using groundtruth object selections. As shown in Fig.~\ref{fig:simulation_evaluation}, our model consistently outperformed \textbf{Binary} at building all four structures. Our model successfully built 58/112 (52\%) more structures that requires modeling complex spatial relations (i.e., circles and table settings) and 14/95 (15\%) structures that can be described by pairwise spatial relations (i.e., lines and towers). In scenes where both methods were successful, our method moved on average $0.17 \pm 0.38$ distractor objects in the scenes while \textbf{Binary} moved $0.21 \pm 0.41$. This result suggests that our method can more effectively reason about the relevance of the objects in the scene, a necessary ability for rearrangements in clutter. 
Figure~1 show our model can generate rearrangements of different structures and size at different parts of the table conditioned on the language instructions.



\subsection{Physical Robot Experiments}
We deploy our system on a Franka Panda Robot with an arm-mounted RGB-D camera to evaluate real-world object manipulation. We generate grasps using the method from~\cite{mousavian2019graspnet} and use RRT-connect~\cite{kuffner2000rrt} for motion planning. 
We visualize successful rearrangements in Fig.~1. 

Failures of the system are driven by oddly perceived objects. When a large portion of an object is occluded, the system is prone to place the object such that it intersects with other objects. Since our work focus on generating 3D structures, we do not use any sophisticated planning method. As a result, motion planning fails sometimes due to unreachable objects. However, the candidate goal scenes generated by our method could be combined with the vision-based planners in~\cite{qureshi2021nerp} to find feasible motion plans.


\section{Conclusions}\label{sec:conclusions}
We presented an learning-based approach for robot planning and manipulation of multi-objects semantic arrangements. Our method leverages a transformer architecture to generate plans as sequence output from an input scene point cloud and language command. Our results show the benefit of our specific architecture over alternative networks.

While our method outperforms the baselines, 
we leave open the problem of operating directly from natural language, instead using structured language to specify parameters of each structure.
We also don't currently address placement in clutter or finding the optimal order of rearranging actions. Instead, we always build in a predefined order. 
In the future, we will incorporate \structformer{} into a full-fledged task-and-motion planner to examine solving rearrangement problems in a variety of environments, not just on tables.


\clearpage
\newpage
\bibliographystyle{IEEEtran}
\bibliography{main}

\begin{thebibliography}{10}
\providecommand{\url}[1]{#1}
\csname url@samestyle\endcsname
\providecommand{\newblock}{\relax}
\providecommand{\bibinfo}[2]{#2}
\providecommand{\BIBentrySTDinterwordspacing}{\spaceskip=0pt\relax}
\providecommand{\BIBentryALTinterwordstretchfactor}{4}
\providecommand{\BIBentryALTinterwordspacing}{\spaceskip=\fontdimen2\font plus
\BIBentryALTinterwordstretchfactor\fontdimen3\font minus
  \fontdimen4\font\relax}
\providecommand{\BIBforeignlanguage}[2]{{%
\expandafter\ifx\csname l@#1\endcsname\relax
\typeout{** WARNING: IEEEtran.bst: No hyphenation pattern has been}%
\typeout{** loaded for the language `#1'. Using the pattern for}%
\typeout{** the default language instead.}%
\else
\language=\csname l@#1\endcsname
\fi
#2}}
\providecommand{\BIBdecl}{\relax}
\BIBdecl

\bibitem{Cakmak2013TowardsAC}
M.~Cakmak and L.~Takayama, ``Towards a comprehensive chore list for domestic
  robots,'' \emph{IEEE International Conference on Human-Robot Interaction
  (HRI)}, pp. 93--94, 2013.

\bibitem{Batra2020RearrangementAC}
D.~Batra, A.~Chang, S.~Chernova, A.~Davison, J.~Deng, V.~Koltun, S.~Levine,
  J.~Malik, I.~Mordatch, R.~Mottaghi, M.~Savva, and H.~Su, ``Rearrangement: A
  challenge for embodied ai,'' \emph{ArXiv}, vol. abs/2011.01975, 2020.

\bibitem{paxton2021predicting}
C.~Paxton, C.~Xie, T.~Hermans, and D.~Fox, ``Predicting stable configurations
  for semantic placement of novel objects,'' in \emph{Conference on Robot
  Learning (CoRL)}, 2021.

\bibitem{Tellex2020RobotsTU}
S.~Tellex, N.~Gopalan, H.~Kress-Gazit, and C.~Matuszek, ``Robots that use
  language,'' \emph{Annual review of control, robotics, and autonomous
  systems}, no.~3, 2020.

\bibitem{krishna2017visual}
R.~Krishna, Y.~Zhu, O.~Groth, J.~Johnson, K.~Hata, J.~Kravitz, S.~Chen,
  Y.~Kalantidis, L.-J. Li, D.~A. Shamma \emph{et~al.}, ``Visual genome:
  Connecting language and vision using crowdsourced dense image annotations,''
  \emph{International journal of computer vision}, vol. 123, no.~1, pp. 32--73,
  2017.

\bibitem{shridhar2020ingress}
M.~Shridhar, D.~Mittal, and D.~Hsu, ``Ingress: Interactive visual grounding of
  referring expressions,'' \emph{The International Journal of Robotics
  Research}, vol.~39, no. 2-3, pp. 217--232, 2020.

\bibitem{kunze2014combining}
L.~Kunze, C.~Burbridge, M.~Alberti, A.~Thippur, J.~Folkesson, P.~Jensfelt, and
  N.~Hawes, ``Combining top-down spatial reasoning and bottom-up object class
  recognition for scene understanding,'' in \emph{IEEE/RSJ International
  Conference on Intelligent Robots and Systems}.\hskip 1em plus 0.5em minus
  0.4em\relax IEEE, 2014, pp. 2910--2915.

\bibitem{gunther2018context}
M.~G{\"u}nther, J.~Ruiz-Sarmiento, C.~Galindo, J.~Gonz{\'a}lez-Jim{\'e}nez, and
  J.~Hertzberg, ``Context-aware 3d object anchoring for mobile robots,''
  \emph{Robotics and Autonomous Systems}, vol. 110, pp. 12--32, 2018.

\bibitem{mees2020learning}
O.~Mees, A.~Emek, J.~Vertens, and W.~Burgard, ``Learning object placements for
  relational instructions by hallucinating scene representations,'' in
  \emph{2020 IEEE International Conference on Robotics and Automation
  (ICRA)}.\hskip 1em plus 0.5em minus 0.4em\relax IEEE, 2020, pp. 94--100.

\bibitem{janner2018representation}
M.~Janner, K.~Narasimhan, and R.~Barzilay, ``Representation learning for
  grounded spatial reasoning,'' \emph{Transactions of the Association for
  Computational Linguistics}, vol.~6, pp. 49--61, 2018.

\bibitem{venkatesh2020spatial}
S.~G. Venkatesh, A.~Biswas, R.~Upadrashta, V.~Srinivasan, P.~Talukdar, and
  B.~Amrutur, ``Spatial reasoning from natural language instructions for robot
  manipulation,'' in \emph{IEEE International Conference on Robotics and
  Automation (ICRA)}, 2021.

\bibitem{kartmann2020representing}
R.~Kartmann, Y.~Zhou, D.~Liu, F.~Paus, and T.~Asfour, ``Representing spatial
  object relations as parametric polar distribution for scene manipulation
  based on verbal commands,'' in \emph{IEEE/RSJ International Conference on
  Intelligent Robots and Systems (IROS)}.\hskip 1em plus 0.5em minus
  0.4em\relax IEEE, 2020, pp. 8373--8380.

\bibitem{yan2020robotic}
F.~Yan, D.~Wang, and H.~He, ``Robotic understanding of spatial relationships
  using neural-logic learning,'' in \emph{IEEE/RSJ International Conference on
  Intelligent Robots and Systems (IROS)}.\hskip 1em plus 0.5em minus
  0.4em\relax IEEE, 2020, pp. 8358--8365.

\bibitem{bisk2018learning}
Y.~Bisk, K.~J. Shih, Y.~Choi, and D.~Marcu, ``Learning interpretable spatial
  operations in a rich 3d blocks world,'' in \emph{Thirty-Second AAAI
  Conference on Artificial Intelligence}, 2018.

\bibitem{zhu2020hierarchical}
Y.~Zhu, J.~Tremblay, S.~Birchfield, and Y.~Zhu, ``Hierarchical planning for
  long-horizon manipulation with geometric and symbolic scene graphs,'' in
  \emph{IEEE International Conference on Robotics and Automation (ICRA)}, 2021.

\bibitem{kase2020transferable}
K.~Kase, C.~Paxton, H.~Mazhar, T.~Ogata, and D.~Fox, ``Transferable task
  execution from pixels through deep planning domain learning,'' in \emph{IEEE
  International Conference on Robotics and Automation (ICRA)}, 2020, pp.
  10\,459--10\,465.

\bibitem{paxton2019prospection}
C.~Paxton, Y.~Bisk, J.~Thomason, A.~Byravan, and D.~Foxl, ``Prospection:
  Interpretable plans from language by predicting the future,'' in \emph{2019
  International Conference on Robotics and Automation (ICRA)}.\hskip 1em plus
  0.5em minus 0.4em\relax IEEE, 2019, pp. 6942--6948.

\bibitem{zeng2018semantic}
Z.~Zeng, Z.~Zhou, Z.~Sui, and O.~C. Jenkins, ``Semantic robot programming for
  goal-directed manipulation in cluttered scenes,'' in \emph{IEEE International
  Conference on Robotics and Automation (ICRA)}.\hskip 1em plus 0.5em minus
  0.4em\relax IEEE, 2018, pp. 7462--7469.

\bibitem{wilson-corl2019-collection-pushing}
\BIBentryALTinterwordspacing
M.~Wilson and T.~Hermans, ``{Learning to Manipulate Object Collections Using
  Grounded State Representations},'' in \emph{Conference on Robot Learning
  (CoRL)}, 2019. [Online]. Available: \url{https://arxiv.org/abs/1909.07876}
\BIBentrySTDinterwordspacing

\bibitem{rosman2011learning}
B.~Rosman and S.~Ramamoorthy, ``Learning spatial relationships between
  objects,'' \emph{The International Journal of Robotics Research}, vol.~30,
  no.~11, pp. 1328--1342, 2011.

\bibitem{fichtl2014learning}
S.~Fichtl, A.~McManus, W.~Mustafa, D.~Kraft, N.~Kr{\"u}ger, and F.~Guerin,
  ``Learning spatial relationships from 3d vision using histograms,'' in
  \emph{IEEE International Conference on Robotics and Automation (ICRA)}.\hskip
  1em plus 0.5em minus 0.4em\relax IEEE, 2014, pp. 501--508.

\bibitem{mees2017metric}
O.~Mees, N.~Abdo, M.~Mazuran, and W.~Burgard, ``Metric learning for
  generalizing spatial relations to new objects,'' in \emph{IEEE/RSJ
  International Conference on Intelligent Robots and Systems (IROS)}.\hskip 1em
  plus 0.5em minus 0.4em\relax IEEE, 2017, pp. 3175--3182.

\bibitem{yuan2021sornet}
W.~Yuan, C.~Paxton, K.~Desingh, and D.~Fox, ``Sornet: Spatial object-centric
  representations for sequential manipulation,'' 2021.

\bibitem{Paul2016EfficientGO}
R.~Paul, J.~Arkin, N.~Roy, and T.~Howard, ``Efficient grounding of abstract
  spatial concepts for natural language interaction with robot manipulators,''
  in \emph{Robotics: Science and Systems}, 2016.

\bibitem{teodorescu2020spatialsim}
L.~Teodorescu, K.~Hofmann, and P.-Y. Oudeyer, ``Spatialsim: Recognizing spatial
  configurations of objects with graph neural networks,'' \emph{arXiv preprint
  arXiv:2004.04546}, 2020.

\bibitem{hristov2020disentangled}
Y.~Hristov, D.~Angelov, M.~Burke, A.~Lascarides, and S.~Ramamoorthy,
  ``Disentangled relational representations for explaining and learning from
  demonstration,'' in \emph{Conference on Robot Learning}.\hskip 1em plus 0.5em
  minus 0.4em\relax PMLR, 2020, pp. 870--884.

\bibitem{johnson2017clevr}
J.~Johnson, B.~Hariharan, L.~Van Der~Maaten, L.~Fei-Fei, C.~Lawrence~Zitnick,
  and R.~Girshick, ``Clevr: A diagnostic dataset for compositional language and
  elementary visual reasoning,'' in \emph{Proceedings of the IEEE conference on
  computer vision and pattern recognition}, 2017, pp. 2901--2910.

\bibitem{yi2018neural}
K.~Yi, J.~Wu, C.~Gan, A.~Torralba, P.~Kohli, and J.~B. Tenenbaum,
  ``Neural-symbolic vqa: Disentangling reasoning from vision and language
  understanding,'' in \emph{Advances in Neural Information Processing Systems},
  2018, pp. 1039--1050.

\bibitem{ding2020object}
D.~Ding, F.~Hill, A.~Santoro, and M.~Botvinick, ``Object-based attention for
  spatio-temporal reasoning: Outperforming neuro-symbolic models with flexible
  distributed architectures,'' \emph{arXiv preprint arXiv:2012.08508}, 2020.

\bibitem{nazarczuk2020shop}
M.~Nazarczuk and K.~Mikolajczyk, ``Shop-vrb: A visual reasoning benchmark for
  object perception,'' in \emph{2020 IEEE International Conference on Robotics
  and Automation (ICRA)}.\hskip 1em plus 0.5em minus 0.4em\relax IEEE, 2020,
  pp. 6898--6904.

\bibitem{girdhar2020cater}
R.~Girdhar and D.~Ramanan, ``{CATER: A diagnostic dataset for Compositional
  Actions and TEmporal Reasoning},'' in \emph{ICLR}, 2020.

\bibitem{hong2021transformation}
X.~Hong, Y.~Lan, L.~Pang, J.~Guo, and X.~Cheng, ``Transformation driven visual
  reasoning,'' in \emph{Proceedings of the IEEE/CVF Conference on Computer
  Vision and Pattern Recognition}, 2021, pp. 6903--6912.

\bibitem{CLEVRER2020ICLR}
K.~Yi, C.~Gan, Y.~Li, P.~Kohli, J.~Wu, A.~Torralba, and J.~B. Tenenbaum,
  ``{CLEVRER:} collision events for video representation and reasoning,'' in
  \emph{ICLR}, 2020.

\bibitem{zhang2019raven}
C.~Zhang, F.~Gao, B.~Jia, Y.~Zhu, and S.-C. Zhu, ``Raven: A dataset for
  relational and analogical visual reasoning,'' in \emph{Proceedings of the
  IEEE/CVF Conference on Computer Vision and Pattern Recognition}, 2019, pp.
  5317--5327.

\bibitem{barrett2018measuring}
D.~Barrett, F.~Hill, A.~Santoro, A.~Morcos, and T.~Lillicrap, ``Measuring
  abstract reasoning in neural networks,'' in \emph{International conference on
  machine learning}.\hskip 1em plus 0.5em minus 0.4em\relax PMLR, 2018, pp.
  511--520.

\bibitem{suhr2017corpus}
A.~Suhr, M.~Lewis, J.~Yeh, and Y.~Artzi, ``A corpus of natural language for
  visual reasoning,'' in \emph{Proceedings of the 55th Annual Meeting of the
  Association for Computational Linguistics (Volume 2: Short Papers)}, 2017,
  pp. 217--223.

\bibitem{achlioptas2018learning}
P.~Achlioptas, O.~Diamanti, I.~Mitliagkas, and L.~Guibas, ``Learning
  representations and generative models for 3d point clouds,'' in
  \emph{International conference on machine learning}.\hskip 1em plus 0.5em
  minus 0.4em\relax PMLR, 2018, pp. 40--49.

\bibitem{park2019deepsdf}
J.~J. Park, P.~Florence, J.~Straub, R.~Newcombe, and S.~Lovegrove, ``Deepsdf:
  Learning continuous signed distance functions for shape representation,'' in
  \emph{Proceedings of the IEEE/CVF Conference on Computer Vision and Pattern
  Recognition}, 2019, pp. 165--174.

\bibitem{mo2019structurenet}
K.~Mo, P.~Guerrero, L.~Yi, H.~Su, P.~Wonka, N.~Mitra, and L.~Guibas,
  ``Structurenet: Hierarchical graph networks for 3d shape generation,''
  \emph{ACM Transactions on Graphics (TOG), Siggraph Asia 2019}, vol.~38,
  no.~6, p. Article 242, 2019.

\bibitem{wu2020pq}
R.~Wu, Y.~Zhuang, K.~Xu, H.~Zhang, and B.~Chen, ``Pq-net: A generative part
  seq2seq network for 3d shapes,'' in \emph{Proceedings of the IEEE/CVF
  Conference on Computer Vision and Pattern Recognition}, 2020, pp. 829--838.

\bibitem{li2017grass}
J.~Li, K.~Xu, S.~Chaudhuri, E.~Yumer, H.~Zhang, and L.~Guibas, ``Grass:
  Generative recursive autoencoders for shape structures,'' \emph{ACM
  Transactions on Graphics (TOG)}, vol.~36, no.~4, pp. 1--14, 2017.

\bibitem{li2019grains}
M.~Li, A.~G. Patil, K.~Xu, S.~Chaudhuri, O.~Khan, A.~Shamir, C.~Tu, B.~Chen,
  D.~Cohen-Or, and H.~Zhang, ``Grains: Generative recursive autoencoders for
  indoor scenes,'' \emph{ACM Transactions on Graphics (TOG)}, vol.~38, no.~2,
  pp. 1--16, 2019.

\bibitem{Chaeibakhsh-icores2021-room-design}
\BIBentryALTinterwordspacing
S.~Chaeibakhsh, R.~S. Novin, T.~Hermans, A.~Merryweather, and A.~Kuntz,
  ``{Optimizing Hospital Room Layout to Reduce the Risk of Patient Falls},'' in
  \emph{{International Conference on Operations Research and Enterprise Systems
  (ICORES)}}, 2021. [Online]. Available: \url{https://arxiv.org/abs/2101.03210}
\BIBentrySTDinterwordspacing

\bibitem{jiang2012learning}
Y.~Jiang, M.~Lim, and A.~Saxena, ``Learning object arrangements in 3d scenes
  using human context,'' in \emph{Proceedings of the 29th International
  Coference on International Conference on Machine Learning}.\hskip 1em plus
  0.5em minus 0.4em\relax Omnipress, 2012, p. 907–914.

\bibitem{vaswani2017attention}
A.~Vaswani, N.~Shazeer, N.~Parmar, J.~Uszkoreit, L.~Jones, A.~N. Gomez,
  {\L}.~Kaiser, and I.~Polosukhin, ``Attention is all you need,'' in
  \emph{Advances in neural information processing systems}, 2017, pp.
  5998--6008.

\bibitem{guo2021pct}
M.-H. Guo, J.-X. Cai, Z.-N. Liu, T.-J. Mu, R.~R. Martin, and S.-M. Hu, ``{PCT}:
  Point cloud transformer,'' \emph{Computational Visual Media}, vol.~7, no.~2,
  pp. 187--199, 2021.

\bibitem{zhou2019continuity}
Y.~Zhou, C.~Barnes, J.~Lu, J.~Yang, and H.~Li, ``On the continuity of rotation
  representations in neural networks,'' in \emph{Proceedings of the IEEE/CVF
  Conference on Computer Vision and Pattern Recognition}, 2019, pp. 5745--5753.

\bibitem{coumans2017pybullet}
E.~Coumans and Y.~Bai, ``Pybullet, a python module for physics simulation in
  robotics, games and machine learning,'' 2017.

\bibitem{morrical2021nvisii}
N.~Morrical, J.~Tremblay, Y.~Lin, S.~Tyree, S.~Birchfield, V.~Pascucci, and
  I.~Wald, ``Nvisii: A scriptable tool for photorealistic image generation,''
  \emph{arXiv preprint arXiv:2105.13962}, 2021.

\bibitem{acronym2020}
C.~Eppner, A.~Mousavian, and D.~Fox, ``{ACRONYM}: A large-scale grasp dataset
  based on simulation,'' in \emph{IEEE Int. Conf. on Robotics and Automation,
  {ICRA}}, 2020.

\bibitem{wang2020sceneformer}
X.~Wang, C.~Yeshwanth, and M.~Nie{\ss}ner, ``Scene{F}ormer: Indoor scene
  generation with transformers,'' \emph{arXiv preprint arXiv:2012.09793}, 2020.

\bibitem{radevski2020decoding}
G.~Radevski, G.~Collell, M.~F. Moens, and T.~Tuytelaars, ``Decoding language
  spatial relations to 2d spatial arrangements,'' in \emph{Proceedings of the
  2020 Conference on Empirical Methods in Natural Language Processing:
  Findings}, 2020, pp. 4549--4560.

\bibitem{mousavian2019graspnet}
A.~Mousavian, C.~Eppner, and D.~Fox, ``{6-{DOF} graspnet: Variational grasp
  generation for object manipulation},'' \emph{International Conference on
  Computer Vision}, pp. 2901--2910, 2019.

\bibitem{kuffner2000rrt}
J.~J. Kuffner and S.~M. LaValle, ``{RRT}-connect: An efficient approach to
  single-query path planning,'' in \emph{Proceedings 2000 ICRA. Millennium
  Conference. IEEE International Conference on Robotics and Automation.
  Symposia Proceedings (Cat. No. 00CH37065)}, vol.~2.\hskip 1em plus 0.5em
  minus 0.4em\relax IEEE, 2000, pp. 995--1001.

\bibitem{qureshi2021nerp}
A.~Qureshi, A.~Mousavian, C.~Paxton, M.~Yip, and D.~Fox, ``Nerp: Neural
  rearrangement planning for unknown objects,'' in \emph{Proceedings of
  Robotics: Science and Systems}, 2021.

\end{thebibliography}

\end{document}